\documentclass[conference]{IEEEtran}

\usepackage{graphicx}
\usepackage{float}
\usepackage{lscape}
\usepackage[lined,ruled,linesnumbered]{algorithm2e}
\usepackage{booktabs}
\usepackage{multirow}
\usepackage{multicol}
\usepackage{paralist}
\usepackage{bm}
\usepackage{epsfig}
\usepackage{graphicx}
\usepackage{times}
\usepackage{mathtools}
\usepackage{amsmath}
\usepackage{amsfonts}
\usepackage[font=small]{caption}
\usepackage{subcaption}
\usepackage{color}
\usepackage{comment}
\usepackage[numbers, sort]{natbib}
\usepackage{url}  
\usepackage[pagebackref=true,breaklinks=true,colorlinks,urlcolor=blue,citecolor=blue,linkcolor=blue,bookmarks=true]{hyperref}
\usepackage{listings}
\usepackage{xspace}
\usepackage[table]{xcolor}
\usepackage{setspace}
\usepackage{nicefrac}
\usepackage{microtype}
\usepackage[utf8]{inputenc} 
\usepackage[T1]{fontenc}    
\renewcommand{\baselinestretch}{1}
\definecolor{LavenderBlue}{rgb}{0.8,0.8,1.0}
\definecolor{Lightapricot}{rgb}{0.99,0.84,0.69}

\newcommand{\heading}[1]{\noindent\textbf{#1}}

\newcommand{\algoNameFull}{Neural Motion Fields\xspace}

\begin{document}

\title{
Neural Motion Fields: Encoding Grasp Trajectories as Implicit Value Functions
}

\author{
Yun-Chun Chen$^{1,2,\ast}$ 
\hspace{2.0mm} 
Adithyavairavan Murali$^{1,\ast}$ 
\hspace{2.0mm} 
Balakumar Sundaralingam$^{1,\ast}$ 
\\
Wei Yang$^1$ 
\hspace{2.0mm} 
Animesh Garg$^{1,2}$ 
\hspace{2.0mm} 
Dieter Fox$^{1,3}$ 
\\
$^1$NVIDIA 
\hspace{2.0mm} 
$^2$University of Toronto 
\hspace{2.0mm} 
$^3$University of Washington 
\hspace{2.0mm} 
$^\ast$Equal contribution
\\
}

\maketitle

\begin{abstract}
The pipeline of current robotic pick-and-place methods typically consists of several stages: grasp pose detection, finding inverse kinematic solutions for the detected poses, planning a collision-free trajectory, and then executing the open-loop trajectory to the grasp pose with a low-level tracking controller. 
While these grasping methods have shown good performance on grasping static objects on a table-top, the problem of grasping dynamic objects in constrained environments remains an open problem. 
We present \algoNameFull, a novel object representation which encodes both object point clouds and the relative task trajectories as an implicit value function parameterized by a neural network. 
This object-centric representation models a continuous distribution over the SE$(3)$ space and allows us to perform grasping reactively by leveraging sampling-based MPC to optimize this value function.
\end{abstract}

\section{Introduction}

Current robotic grasping approaches typically decompose the task of grasping into several sub-components: detecting grasp poses on point clouds~\cite{ContactGraspNet, GADDPG}, finding inverse kinematics solutions at these poses, solving collision-free trajectories to pre-grasp standoff poses and finally executing open-loop trajectories from standoff poses to the grasp poses \cite{GraspNet, CollisionNet}. 
By inferring a finite discrete number of grasp poses, such an approach neglects the insight that object grasp affordances are a continuous manifold. While this approach has yielded tremendous progress in bin-picking and grasping unknown objects on a table-top, reactive grasping of unknown objects in constrained environments remains an open problem. 

\begin{figure}[t]
  \centering
  \includegraphics[width=0.9\linewidth]{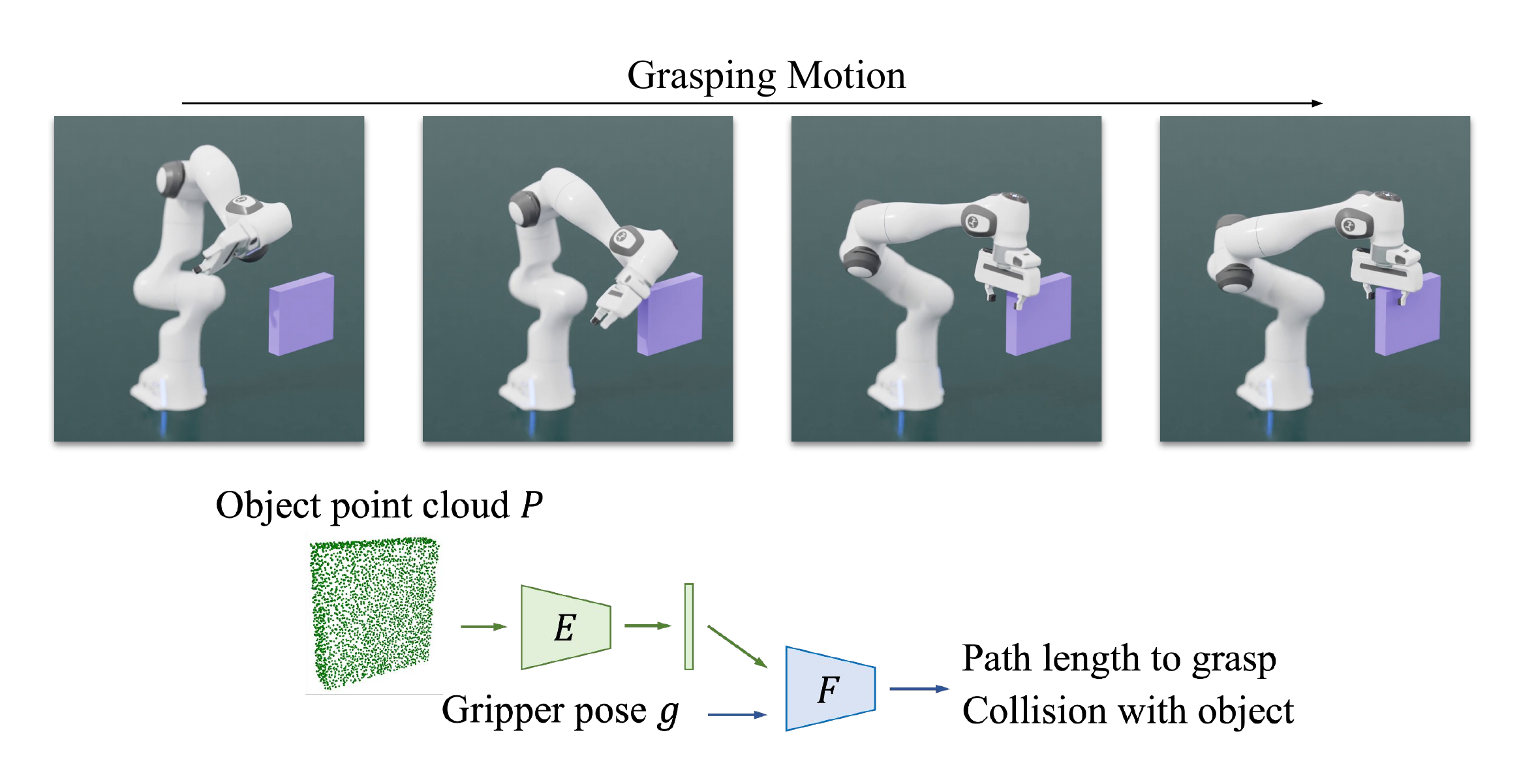}
  \caption{
  We present \algoNameFull that learns a value function that can be queried to generate grasping motion.
  We use separate network weights for predicting path lengths and collisions.
  }
  \label{fig:intro}
  \vspace{-4.0mm}
\end{figure}

Implicit neural representations~\cite{NeRF, DeepSDF, mescheder2019occupancy} have emerged as a new paradigm for applications in rendering, view synthesis, and shape reconstruction. 
Compared to traditional explicit representations (e.g., point clouds and meshes), implicit neural representations can represent continuous signals at arbitrarily high resolutions. 
Motivated by implicit neural representations, we propose to learn a value function that encodes robotic task trajectories using a neural network.
Our key insight is to map each gripper pose in SE$(3)$ to its trajectory path length as shown in Figure~\ref{fig:intro}.
To train the model, we generate synthetic data of grasping process by using a prior grasp dataset~\cite{GraspNet} and planning trajectories with a RRT~\cite{RRT} motion planner.
Once the model training is done, we cast the learned value function as a cost and leverage the Model Predictive Control (MPC)~\cite{bhardwaj2022storm} algorithm to query gripper poses in SE(3) with cost minimization.
This allows us to generate a kinematically feasible trajectory that the robot can execute to reach a grasp on the object. 

We benchmark our method on the grasping task and report the success rate.
In addition, we evaluate our model in two settings: static object poses and dynamic object poses, and provide ablation studies under various settings. 
In summary, our contributions include
\begin{enumerate}
  \item We propose \algoNameFull, a novel formulation of the grasp motion generation problem in SE$(3)$ as a continuous implicit representation.
  \item We show that this learned object-centric representations allows reactive grasp manipulation using MPC~\cite{bhardwaj2022storm}.
\end{enumerate}

\section{Problem Statement}

\heading{Value function learning for grasping.}
We are interested in learning a model that can be used to plan a kinematically feasible trajectory for the robot to execute to grasp an object. 
Specifically, we cast this task as a value function learning problem. 
We assume that we are given a segmented object point cloud $P \in \mathbb{R}^{N \times 3}$, where $N$ is the number of points in a point cloud, and a gripper pose $g \in \mathrm{SE}(3)$.
The value function $V(g, P)$ describes how far the gripper pose $g$ is from a grasp on the object.
We use the path length of a gripper pose to represent the value function.

\begin{figure}[t]
  \centering
  \begin{subfigure}[b]{0.49\linewidth}
    \includegraphics[width=\linewidth]{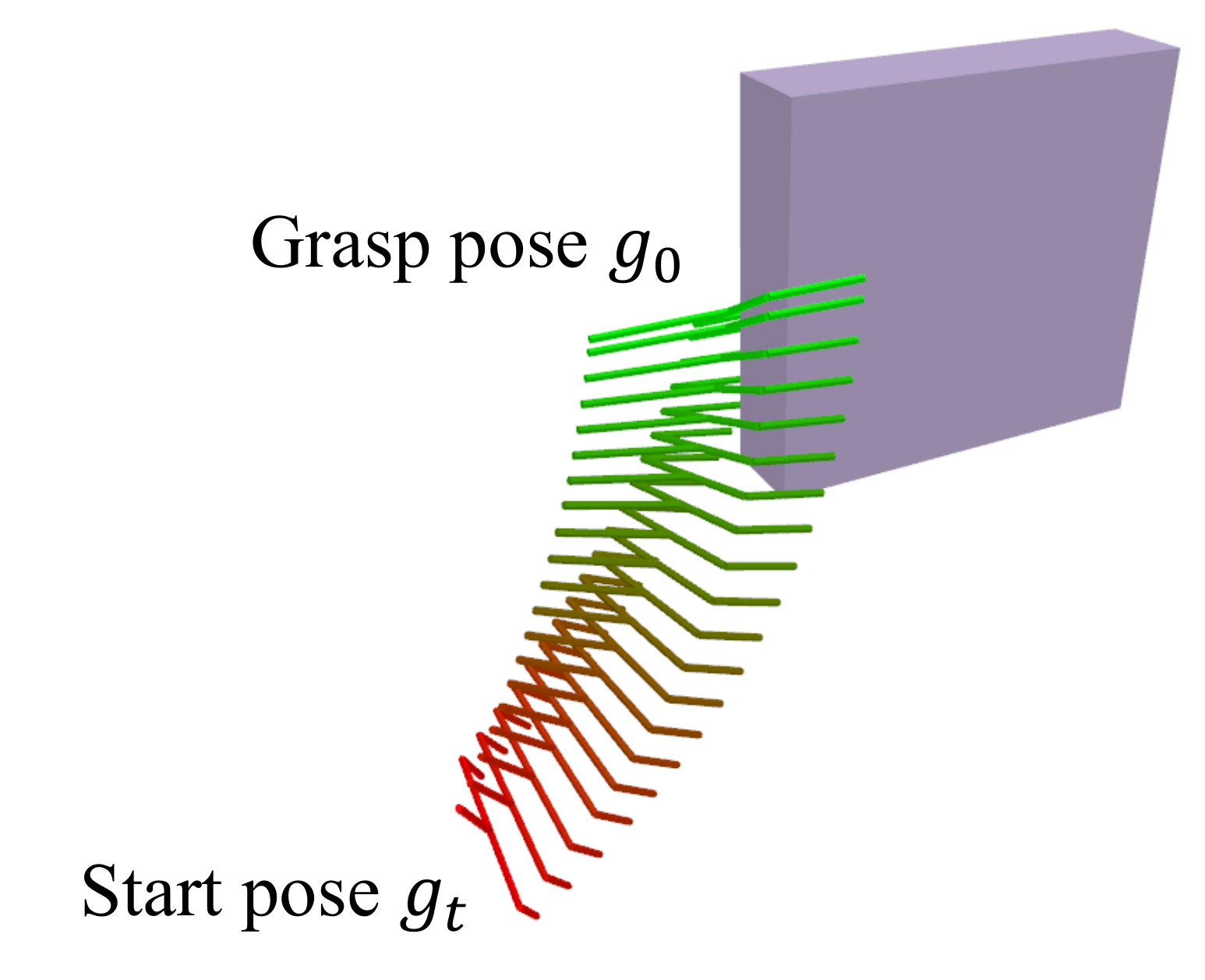}
    \vspace{-4.0mm}
    \caption{An example trajectory.}
    \label{fig:trajectory}
  \end{subfigure}
  \hfill
  \begin{subfigure}[b]{0.49\linewidth}
    \includegraphics[width=\linewidth]{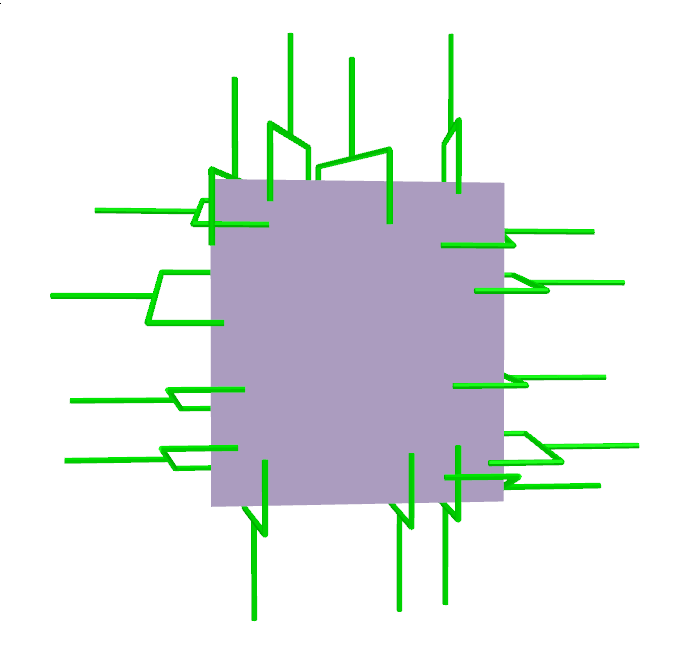}
    \vspace{-4.0mm}
    \caption{Anchor grasps.}
    \label{fig:anchor-grasps}
  \end{subfigure}
  \caption{
  (a) An example trajectory planned by RRT. 
  Red represents longer path length.
  Green represents shorter path length.
  (b) Visualization of the selected anchor grasps.
  }
  \label{fig:path-length}
  \vspace{-6.0mm}
\end{figure}

\vspace{1.5mm}
\heading{Gripper pose path length.}
As shown in Figure~\ref{fig:trajectory}, given a trajectory $\{g_i\}_{i=0}^t$, where $g_0$ denotes the end pose (grasp pose) and $g_t$ denotes the start pose, the path length of the start pose $g_t$ (denoted as $V(g_t)$) is defined as the cumulative sum of the average distance between two adjacent gripper poses~\cite{PoseCNN}, which can be expressed by 
\begin{equation}
  V(g_t) = \sum_{i=0}^{t-1} \frac{1}{m}\sum_{x \in M}^{} \| (R_i x + T_i) - (R_{i+1} x + T_{i+1})\|,
\end{equation}
where $R_i$ is the rotation of the gripper pose $g_i$, $T_i$ is the translation of the gripper pose $g_i$, $m$ is the number of keypoints of the gripper, and $M$ is the set of keypoints of the gripper.

\section{\algoNameFull}

\subsection{Learning from Grasp Trajectories}

Given an object point cloud $P$ and a gripper pose $g$, our goal is to learn a model that approximates the value function $V(g, P)$.
We propose \algoNameFull, which consists of two modules: a path length module and a collision module.

\vspace{1.5mm}
\heading{Path length prediction.}
As shown in Figure~\ref{fig:intro}, the path length module first takes as input the object point cloud $P$ and uses the point cloud encoder $E_\text{path-length}$ to encode a feature embedding $f_\text{path-length} = E_\text{path-length}(P) \in \mathbb{R}^{d}$, where $d$ is the dimension of the feature $f_\text{path-length}$.
Then, the point cloud feature $f_\text{path-length}$ and the gripper pose $g$ are concatenated and passed to the path length prediction network $F_\text{path-length}$ to predict the path length $V_\text{pred}(g, P)$ for the gripper pose $g$.\footnote{The gripper pose $g$ is first converted to a vector in $\mathbb{R}^9$ which is the contatenatation of the 6D rotation representation~\cite{zhou2019continuity} of $g$ and the translation vector in $\mathbb{R}^3$ of $g$. The 9D vector will then be concatenated with the point cloud feature $f_\text{path-length}$ for path length prediction.}

To train the path length module, we adopt an $\ell_1$ loss function, which is defined as
\begin{equation}
  \mathcal{L}_\text{path-length} = \|V_\text{pred}(g, P) - V_\text{gt}(g, P)\|_1,
\end{equation}
where $V_\text{pred}(g, P)$ denotes the predicted path length of the gripper pose $g$ and $V_\text{gt}(g, P)$ denotes the ground truth.

\begin{figure}[t]
  \centering
  \includegraphics[width=0.9\linewidth]{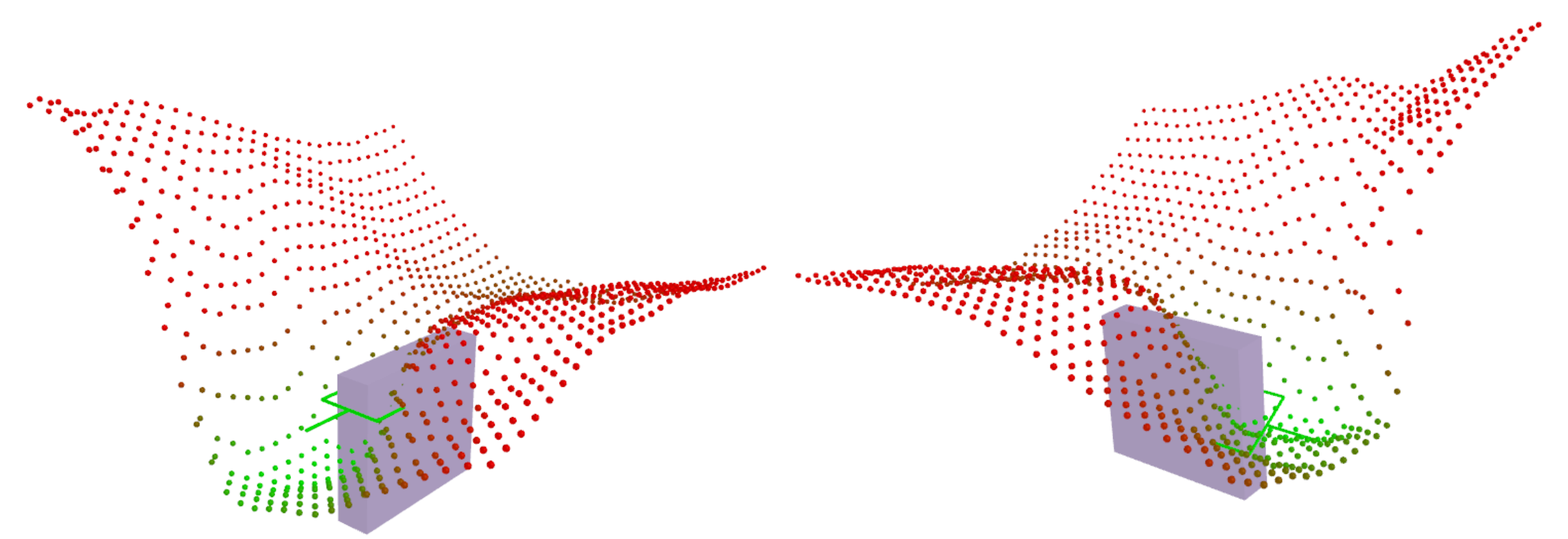}
  \caption{
  We visualize two cost maps of a box object.
  In each cost map, we select a grasp pose.
  We keep the orientation, vary the x and y positions, and query the path length of the composed gripper pose using our model.
  Each gripper pose is represented by a point in 3D.
  Red represents longer path length (higher in z).
  Green represents shorter path length (lower in z).
  }
  \label{fig:cost-map}
  \vspace{-4.0mm}
\end{figure}

We visualize the learned value function using a cost map visualization as shown in Figure~\ref{fig:cost-map}.
We show two cost maps.
In each cost map, we select a grasp pose.
We keep the orientation and vary the x and y positions of the grasp pose to compose poses.
We then query the path lengths of the composed poses using the learned model.
Each input pose is represented by a point in $\mathbb{R}^3$ and is colored by its predicted path length (red means longer path length, while green means shorter).

\vspace{1.5mm}
\heading{Collision prediction.}
Having the path length module alone is insufficient as the model does not explicitly penalize collisions between the gripper and the object of interest.
To address this issue, we develop a collision module as shown in Figure~\ref{fig:intro} (same input as the path length module, but mapped to a different output using a different set of network weights).

Given an object point cloud $P$ and a gripper pose $g$, the collision module first uses the point cloud encoder $E_\text{collision}$ to encode the point cloud feature $f_\text{collision}~=~E_\text{collision}(P)~\in~\mathbb{R}^{k}$, where $k$ is the dimension of the feature $f_\text{collision}$.
Then, the point cloud feature $f_\text{collision}$ and the gripper pose $g$ are concatenated and passed to the collision prediction network $F_\text{collision}$ to predict the probability $p_\text{pred}(g, P)$ of the gripper pose $g$ being in collision with the object.

To train the collision model, we adopt a standard binary cross-entropy loss function, which is defined as
\begin{equation}
  \begin{split}
    \mathcal{L}_\text{collision} & = p_\text{gt}(g, P) \log p_\text{pred}(g, P) \\
    & + (1-p_\text{gt}(g, P)) \log (1 - p_\text{pred}(g, P)),
  \end{split}
\end{equation}
where $p_\text{pred}(g, P)$ is the predicted collision probability and $p_\text{gt}(g, P)$ is the ground truth.

\subsection{Generating Grasp Motion}

Given the path length value function represented by the path length module and the collision value function represented by the collision module, we formulate the grasp cost~$\mathcal{C}_\text{grasp}$ as
\begin{equation}
  \mathcal{C}_\text{grasp}(g_t, P) = (1 - V(g_t, P)) + C(g_t, P),
\end{equation}
where $V(g_t, P)$ is the predicted path length for the gripper pose $g_t$, $C(g_t, P)$ is the collision cost of the gripper pose $g_t$ computed by thresholding $p(g_t, P)\geq\tau$, $\tau$ is a hyperparameter, and $P$ is the object point cloud. 
In our work, we set $\tau = 0.25$.

We then optimize the grasp cost $\mathcal{C}_\text{grasp}$ along with the cost~$\mathcal{C}_\text{storm}$ to ensure smooth collision-free motions using STORM~\cite{bhardwaj2022storm}, which is a GPU-based MPC framework:
\begin{equation}
    \min_{\ddot{x}_{t\in[0,H]}} \quad \mathcal{C}_{\text{storm}}(q) + \mathcal{C}_{\text{grasp}}
    \label{eq:grasp-cost}
\end{equation}
Additional details on~$\mathcal{C}_{\text{storm}}$ is available in~\cite{bhardwaj2022storm}.

\begin{figure*}[t]
  \centering
  \begin{subfigure}[t]{0.48\linewidth}
    \includegraphics[width=0.49\linewidth]{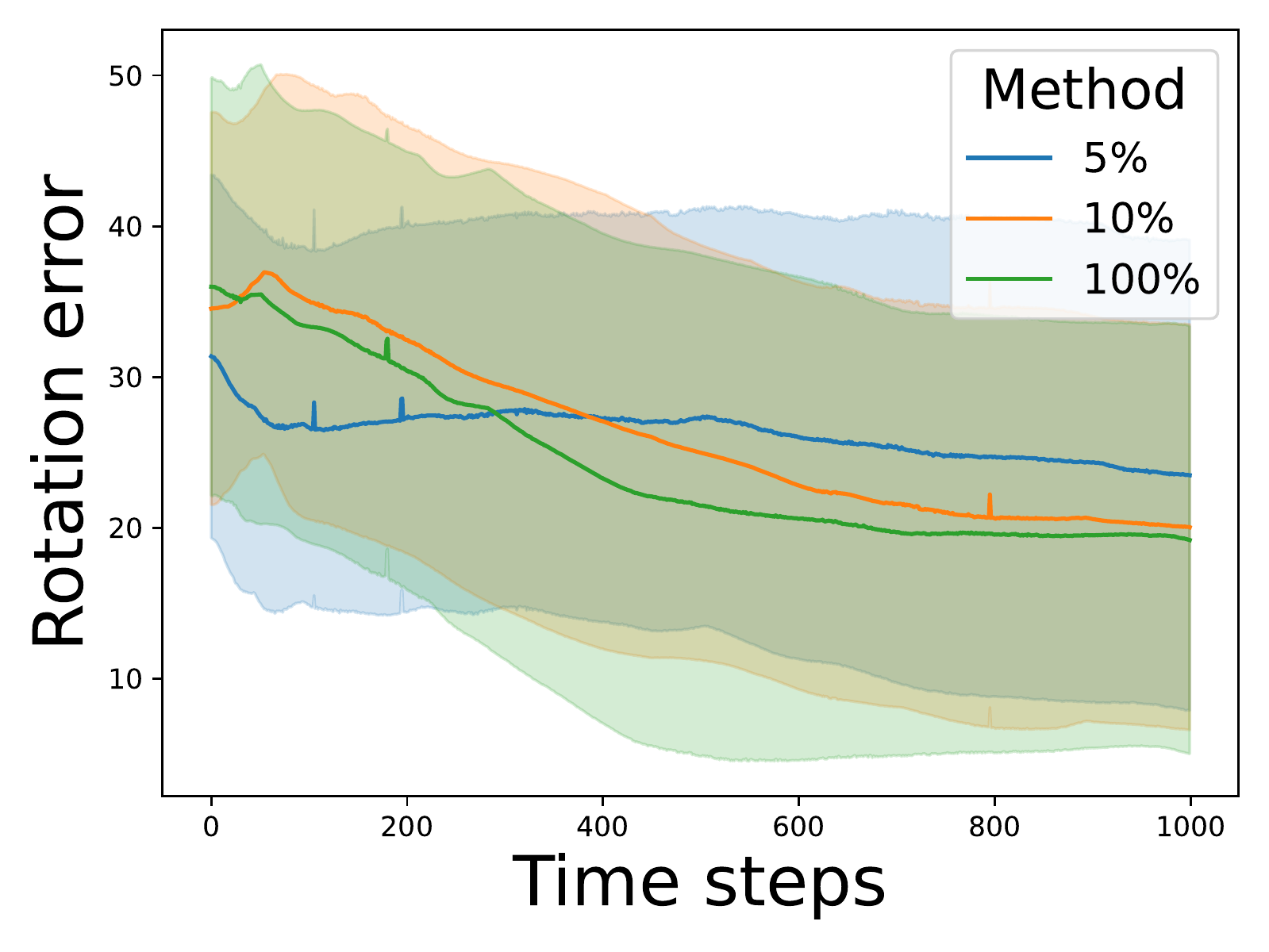}
    \includegraphics[width=0.49\linewidth]{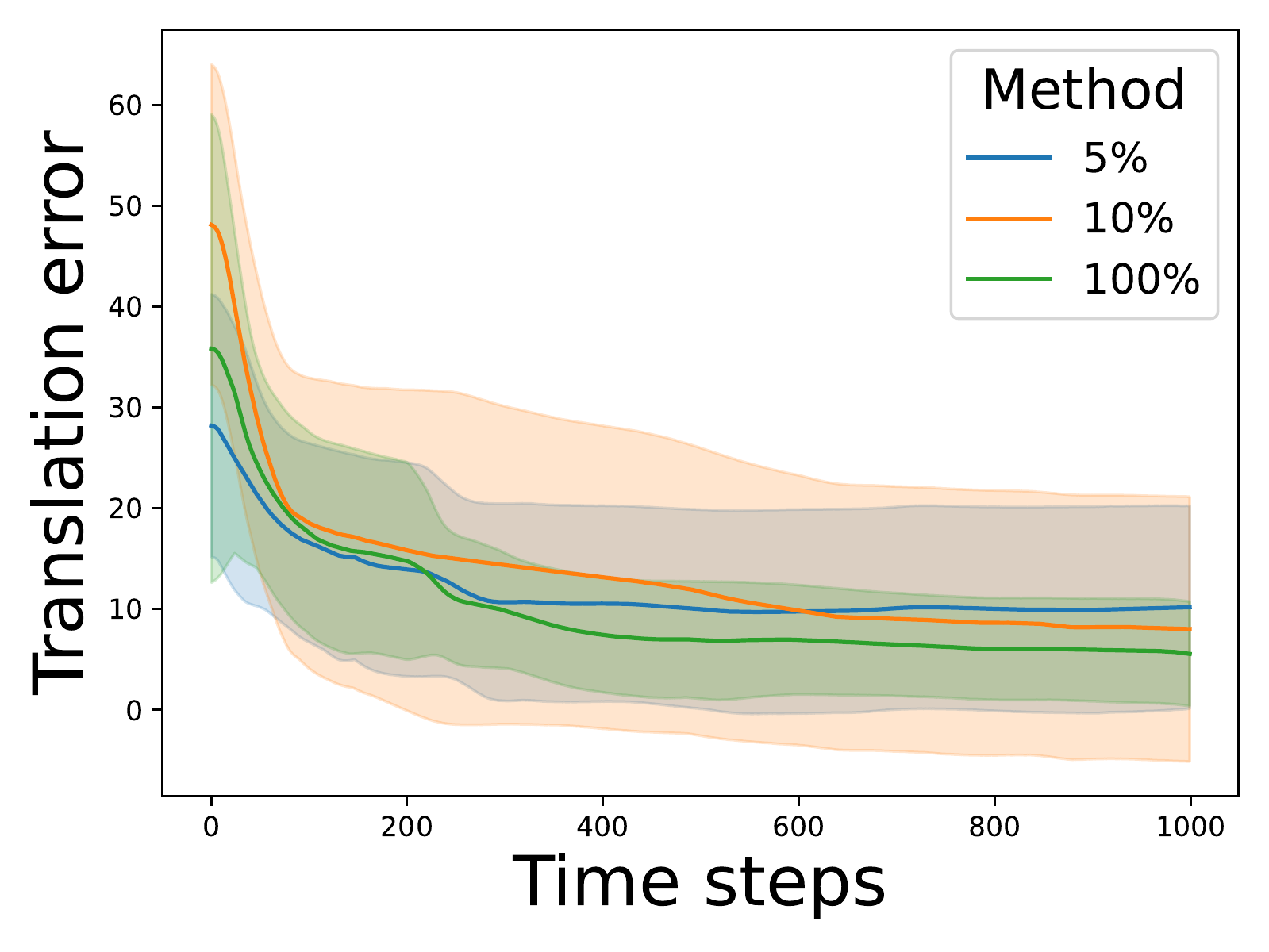}
    \vspace{-1.0mm}
    \caption*{Static Object Poses}
  \end{subfigure}
  \hfill
  \begin{subfigure}[t]{0.48\linewidth}
    \includegraphics[width=0.49\linewidth]{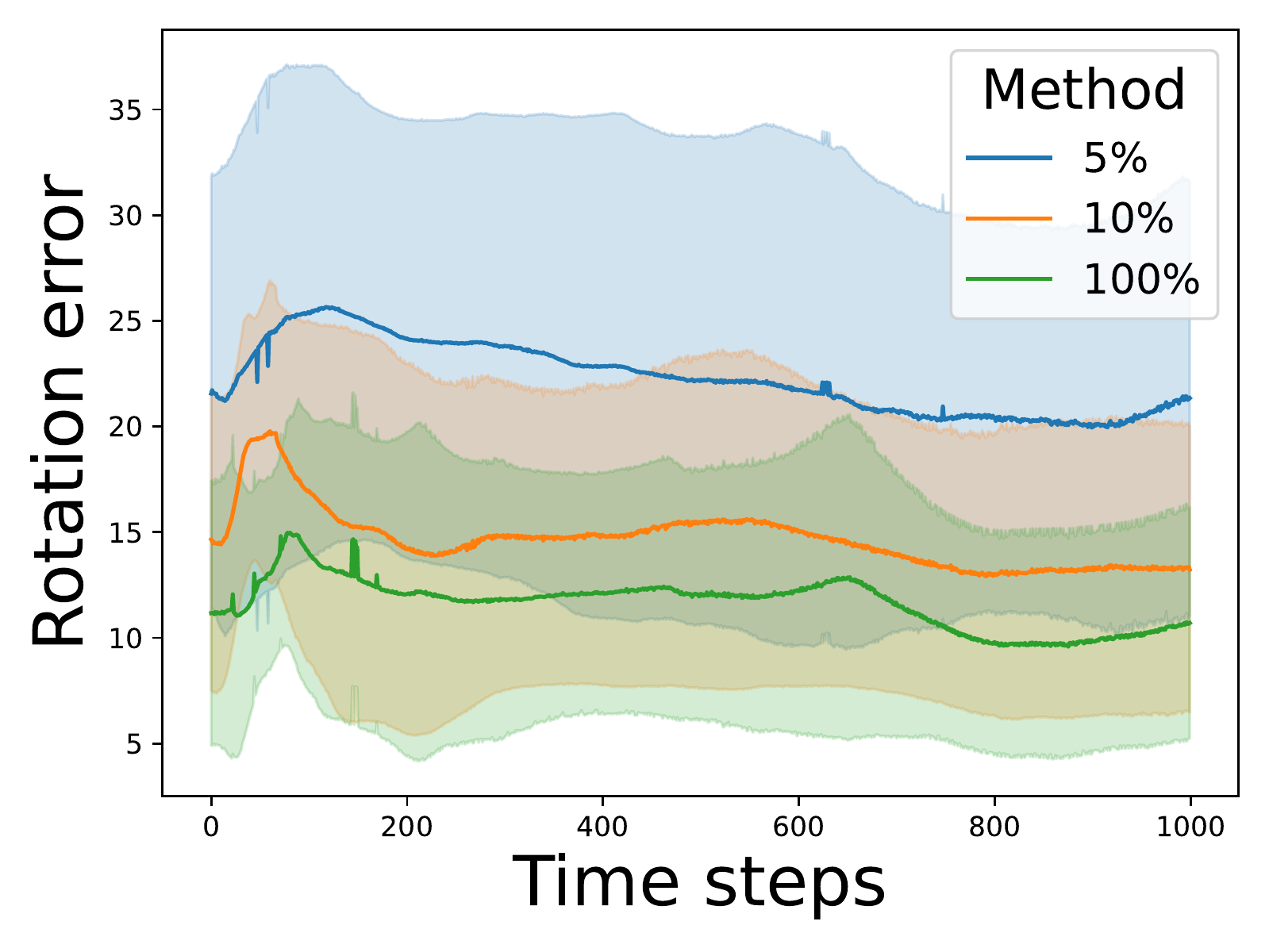}
    \includegraphics[width=0.49\linewidth]{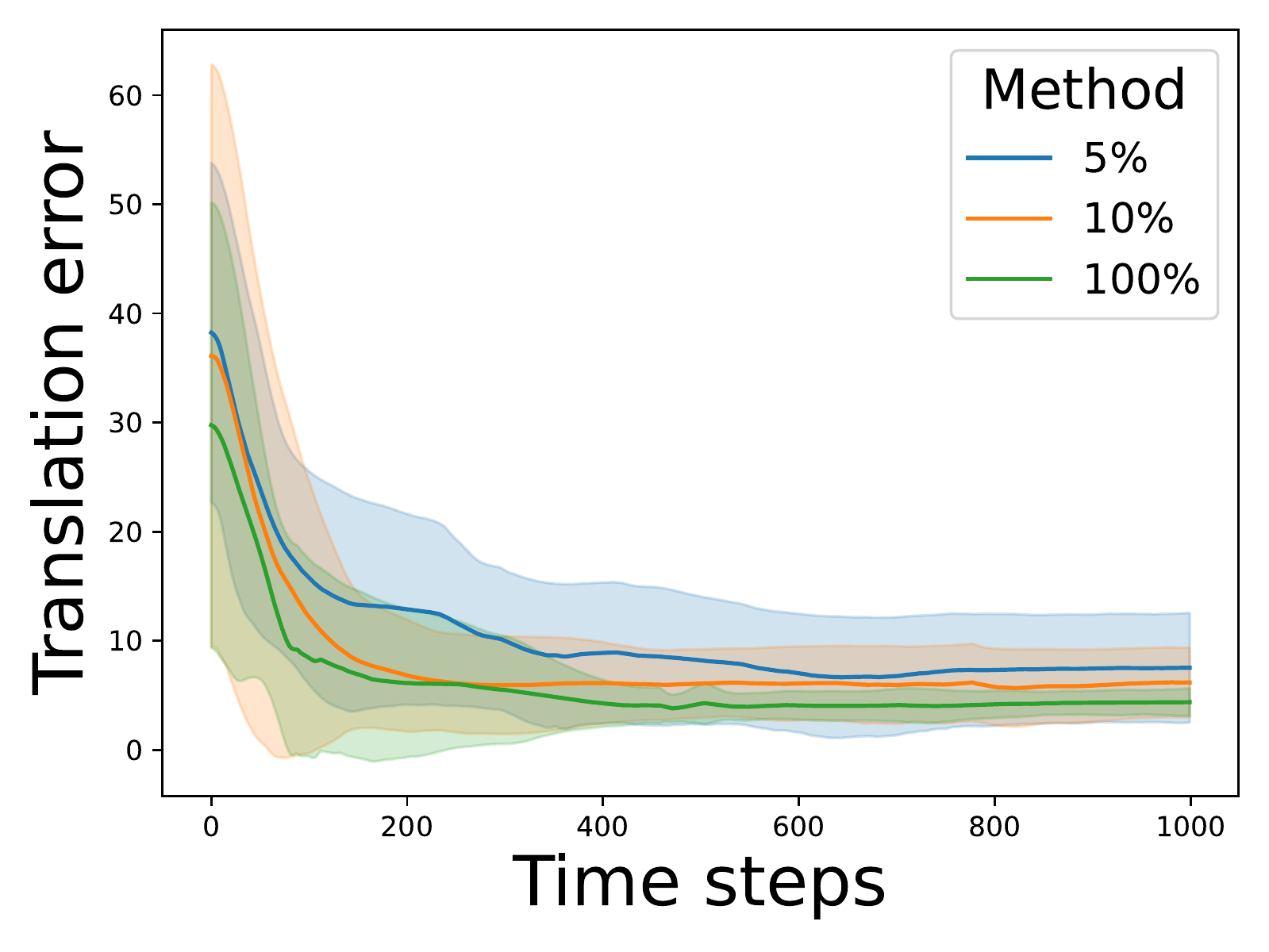}
    \vspace{-1.0mm}
    \caption*{Dynamic Object Poses}
  \end{subfigure}
  \caption{
  \textbf{Ablation study on the number of trajectories.}
  We show the rollout curves of our model trained on differnt numbers of trajectories. 
  }
  \label{fig:dataset-ablation}
  \vspace{-3.0mm}
\end{figure*}

\section{Experiments}

\subsection{Experimental Setup}

\heading{Dataset.}
We experiment with four box objects from the dataset provided by \cite{GraspNet} and one bowl object from the ACRONYM dataset~\cite{eppner2021acronym}.
We first subsample a set of 16 grasp poses around the object using farthest point sampling as shown in Figure~\ref{fig:anchor-grasps}. 
We then apply a 180 degree rotation along the z-axis of the gripper pose to get the flipped counterparts.
The selected 32 grasps are then used as the goal poses for trajectory data collection.
We randomly sample a gripper pose with respect to one of the goal poses and use the RRT~\cite{RRT} planner from OMPL~\cite{sucan2012the-open-motion-planning-library} to plan a trajectory between the two poses.
We note that our data generation pipeline is agnostic to the choice of the motion planning algorithm and other planners are also applicable.
We use inverse kinematics to solve for the joint angles for each waypoint returned by RRT and interpolate between two adjacent joint angles to obtain denser waypoints for each trajectory. 
For each object, we collect one million trajectories. 
We further filter out trajectories where the start gripper pose has a path length greater than a threshold $\phi = 1.0$.

\vspace{1.5mm}
\heading{Implementation details.}
We implement our model using PyTorch~\cite{PyTorch}.
We use the ADAM~\cite{ADAM} optimizer for model training.
We use DGCNN~\cite{DGCNN} to be our point cloud encoder.
The path length prediction network $F_\text{path-length}$ and the collision prediction network $F_\text{collision}$ both consist of 20 fully connected layers.
The learning rate is set to $2 \times 10^{-3}$ with a weight decay of $1 \times 10^{-6}$.
We train our model using eight NVIDIA V100 GPUs with 32GB memory each.
The batch size is set to 32.
The number of points in a point cloud is set to 1,024.
The dimensions for the point cloud features $f_\text{path-length}$ and $f_\text{collision}$ are both 512.
The training time for both the path length module and the collision module is around 7 days.

\begin{table}[t]
  \small
  \begin{center}
  \caption{
  Grasp success rate.
  }
  \vspace{-1.5mm}
  \label{table:grasping-exp}
  {
  \addtolength{\tabcolsep}{-3.1pt}
  \begin{tabular}{lccccc}
    \toprule
    \rowcolor{LavenderBlue}
    Method & Bowl & Box A & Box B & Box C & Box D \\
    \midrule
    Oracle & 40\% & 100\% & 40\% & 30\% & 50\% \\
    Ours & 30\% & 80\% & 40\% & 40\% & 30\% \\
    \bottomrule
  \end{tabular}
  }
  \end{center}
  \vspace{-5.0mm}
\end{table}

\subsection{Evaluation on Grasping}

\heading{Setting.}
For each object, there are 10 test cases.
In each test case, we initialize the object with a stable pose on the tabletop.
In each test case, we optimize over the learned path length module and the collision module to find the minimum point by leveraging sampling based optimization~\cite{bhardwaj2022storm}.
The optimized gripper pose is then used as the grasp pose.
We use STORM pose reaching~\cite{bhardwaj2022storm} to reach the grasp pose.
We set the time limit for each test case to 30 seconds.

\vspace{1.5mm}
\heading{Metric.}
We use the grasp success rate to evaluate performance.

\vspace{1.5mm}
\heading{Results.}
Table~\ref{table:grasping-exp} reports the grasp success rate of the 5 objects.
Our model performs well on Box A, but achieves inferior performance on all other objects.
The inferior performance is due to the collision between the finger tips of the gripper and the object.

\begin{figure}[t]
  \centering
  \begin{subfigure}[t]{0.48\linewidth}
    \includegraphics[width=\linewidth]{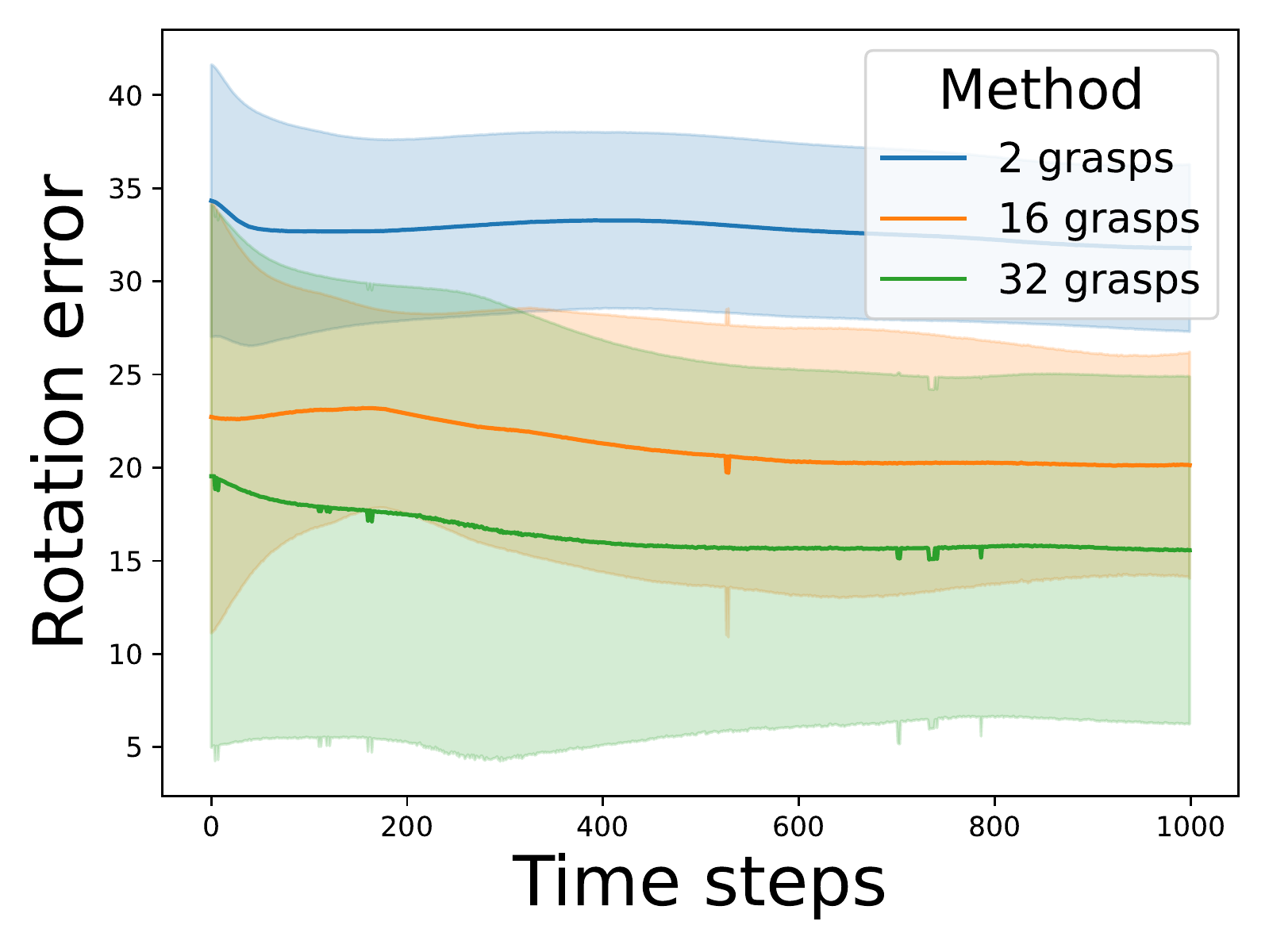}
  \end{subfigure}
  \hfill
  \begin{subfigure}[t]{0.48\linewidth}
    \includegraphics[width=\linewidth]{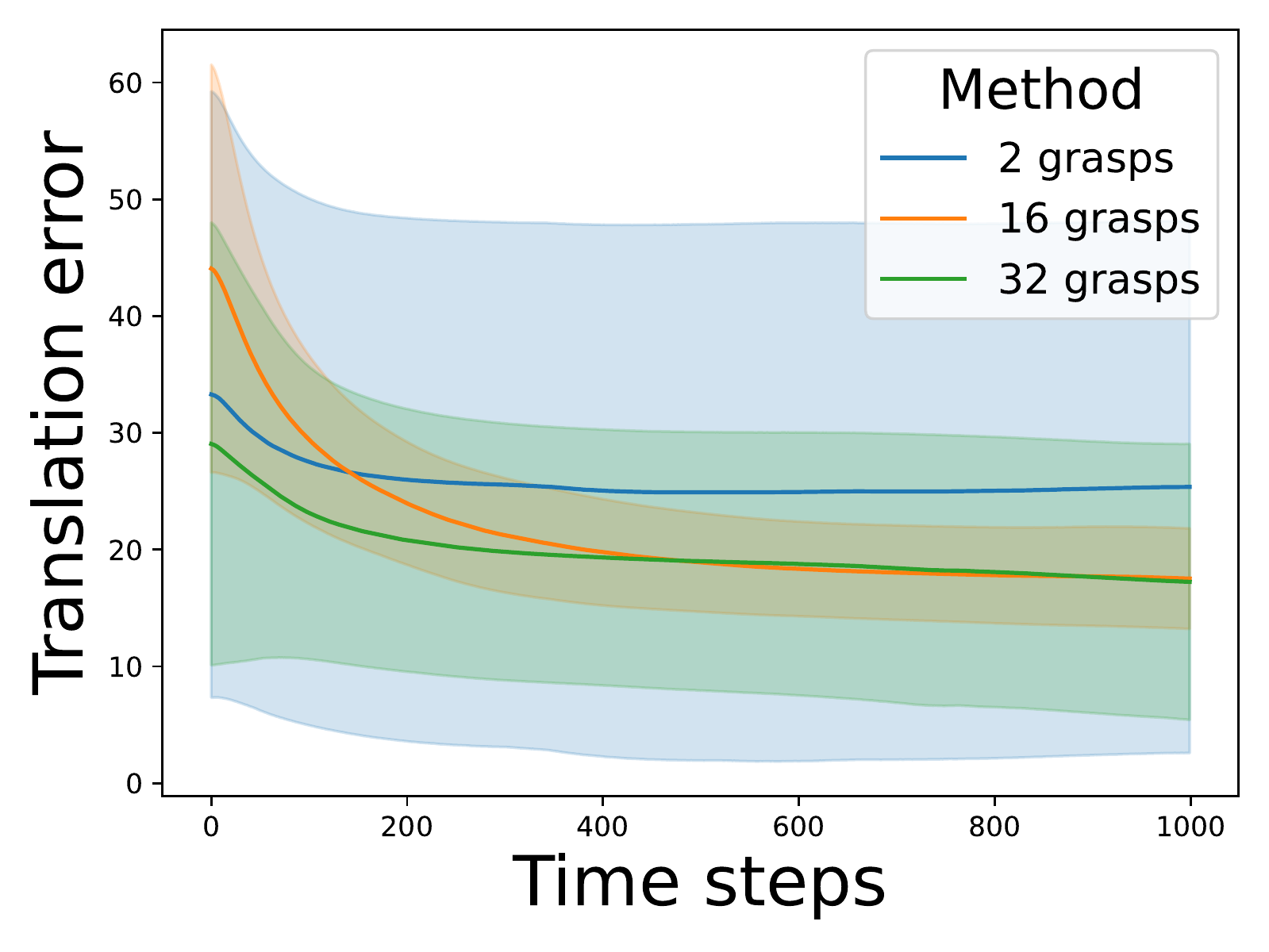}
  \end{subfigure}
  \caption{
  \textbf{Ablation study on the number of anchor grasps.}
  We show the rollout curves of our model trained on trajectories collected by using differnt numbers of anchor grasps. 
  }
  \label{fig:num-grasps}
  \vspace{-4.0mm}
\end{figure}

\subsection{Ablation Study}

\heading{Setting.}
We conduct two ablation studies: 1) static object pose reaching and 2) dynamic object pose reaching.
In the static object pose reaching setting, we randomly sample an object pose at the beginning of each episode and place the object at the sampled pose.
The object pose is kept static throughout the episode.
In the dynamic object pose reaching setting, we randomly sample an object pose and a velocity vector at the beginning of each episode and place the object at the sampled pose while the object is moving at the speed specified by the velocity vector.
In both settings, there are 50 episodes.
Each episode has 1,000 time steps.
We run MPC with our model to optimize the grasp cost in Equation~\eqref{eq:grasp-cost} and execute the robot.

\vspace{1.5mm}
\heading{Metric.}
We follow \cite{cruciani2020benchmarking} and compute the rotation error and the translation error between the current gripper pose and the closest grasp pose at each time step.
The closest grasp pose is the grasp pose that has the minimum path length from the current gripper pose.

\vspace{1.5mm}
\heading{Ablation study on the number of trajectories.}
Figure~\ref{fig:dataset-ablation} shows the rollout curves of our model trained on different percentages of the collected dataset (i.e., 5\%, 10\% and 100\%).
We observe that in both settings, training on the entire dataset (i.e., 100\%) achieves the best in both the rotation error and the translation error.

\vspace{1.5mm}
\heading{Abaltion study on the number of grasp poses.}
Figure~\ref{fig:num-grasps} shows the rollout curves of our model trained on trajectories collected by using different numbers of anchor grasps (i.e., 2, 16 and 32 grasps) in the dynamic object pose reaching setting.
We observe that the model trained on trajectories generated from 32 anchor grasps achieves the best performance and converges faster in translation error than all other models.

\section{Conclusions and Future Work}

We propose \algoNameFull, a novel object representation which encodes both object point clouds and the relative task trajectories as an implicit value function parameterized by a neural network. 
This object-centric representation models a continuous distribution over the SE$(3)$ space and allows us to perform grasping reactively by leveraging sampling-based MPC to optimize this value function.
Through experimental evaluations, we show that by training on more numbers of anchor grasps and larger scale datasets results in superior performance.
In future work, we plan to train a single model on more objects.

\IEEEpeerreviewmaketitle

\bibliographystyle{plainnat}
\bibliography{references}

\end{document}